\documentclass[a4paper, 10 pt, conference]{ieeeconf}  % Comment this line out if you need a4paper

\IEEEoverridecommandlockouts                              % This command is only needed if 
\usepackage{cite}
\usepackage{comment}
\usepackage{amsmath,amssymb,amsfonts}
\usepackage{algorithm}
\usepackage{algpseudocode}
\usepackage{graphicx}
\usepackage{textcomp}
\usepackage{stfloats}
\usepackage{xcolor}
\def\BibTeX{{\rm B\kern-.05em{\sc i\kern-.025em b}\kern-.08em
    T\kern-.1667em\lower.7ex\hbox{E}\kern-.125emX}}
\begin{document}
\title{Edge Computing Architectures for Enabling the Realisation of the Next Generation Robotic Systems}
\author{Achilleas Santi Seisa$^{1}$, Gerasimos Damigos$^{2}$, Sumeet Gajanan Satpute$^{1}$, Anton Koval$^{1}$ and \\ George Nikolakopoulos$^{1}$%
\thanks{This work has been partially funded by the European Unions Horizon 2020 Research and Innovation Programme AERO-TRAIN under the Grant Agreement No. 953454.}
\thanks{$^{1}$ The authors are with the Robotics and AI Team, Department of Computer, Electrical and Space Engineering, Lule\aa\,\, University of Technology, Lule\aa\,\,}
\thanks{$^{2}$ The author is with Ericsson Reasearch, Lule\aa\,\,}
\thanks{Corresponding Author's email: {\tt\small achsei@ltu.se}}
}
\maketitle
%%%%%%%%%%%%%%%%%%%%%%%%%%%%%%%%%%%%%%%%%%%%%%%%%%%%%%%%%%%%%%%
\begin{abstract}
Edge Computing is a promising technology to provide new capabilities in technological fields that require instantaneous data processing. Researchers in areas such as machine and deep learning use extensively edge and cloud computing for their applications, mainly due to the significant computational and storage resources that they provide. Currently, Robotics is seeking to take advantage of these capabilities as well, and with the development of 5G networks, some existing limitations in the field can be overcome. In this context, it is important to know how to utilize the emerging edge architectures, what types of edge architectures and platforms exist today and which of them can and should be used based on each robotic application. In general, Edge platforms can be implemented and used differently, especially since there are several providers offering more or less the same set of services with some essential differences. Thus, this study addresses these discussions for those who work in the development of the next generation robotic systems and will help to understand the advantages and disadvantages of each edge computing architecture in order to choose wisely the right one for each application. 
\end{abstract}

%%%%%%%%%%%%%%%%%%%%%%%%%%%%%%%%%%%%%%%%%%%%%%%%%%%%%%%%%%%%%%%
\section{Introduction}
\label{intro}
Currently, the robotics community is pushing the autonomy levels for the next generation of applications with autonomous robots~\cite{nikolakopoulos2021pushing}. Among the typical examples, one can highlight the multi-session simultaneous localization and mapping (SLAM), with multiple robots in exploration or map-based navigation missions that include collaborative online point cloud map merging and at the same time provide estimation of robots' localization \cite{agha2021nebula}. These tasks are computationally expensive and are demanding an extensive amount of memory and computational resources. As a response to these demands, the edge computing has a high potential to solve these tasks by offloading them from the on-board computers and support faster executions by providing capacity for real-time processing. Additionally, the integration of the edge architectures with the also emerging 5G communication technologies with gigabit speeds and low latencies will allow to establish a ubiquitous communication between robots~\cite{taleb2017multi, zeng2019accessing}. A typical example of such a demanding application is when solving the multi-robot SLAM problem that is also critical for the robots' risk-aware navigation~\cite{lindqvist2021reactive,lindqvist2021collision} in dynamically changing environments that demands low latencies, as well as a minimal lag in the response to high-level control commands, efficient visual representation, security and privacy of data \cite{nikolakopoulos2008experimental}. 

Towards these challenges, it is expected that the development and utilization of edge computing and 5G technologies can drastically boost the levels of robotic autonomy and result in overcoming existing on-board hardware limitations, which will foster answering these challenges. Alike on-board computers, edge platforms can handle huge amount of data at a very high processing speed, provide strong computation capacity and real-time data transmission. In comparison to cloud computing, edge computing has the capacity to enable local computing, thus dealing more efficiently with big volumes and velocities of generated local data \cite{jararweh2016future}. Moreover, it will provide lower latencies due to the significantly smaller distance between the edge connected robots and the edge datacenters. Thus, in this article, we analyze different types of cloud architectures with the focus on edge-based architectures, especially for robotic systems. 

Edge computing is a promising technology that may foster solving some of the existing fundamental challenges in the field of robotics that will allow to push the realisation of the next generation of robotic systems with higher levels of autonomy and longer operational times. In order for the robotic platforms to operate for a longer period of time, they need to be lighter and have a lower computational load locally, while being equipped with sensors that are capable to generate the necessary data required for low-level autonomy and mission execution. Additionally, in the collaborative missions with the multi-agent systems it is crucial that in some tasks all the robots are able to communicate with each other and exchange data in real-time. At the same time, some processes will require information that was collaboratively generated by all robots \cite{agha2021nebula}. 

With the utilization of edge computing technologies, data that are generated from on-board robotic sensors can be sent to the edge allowing to offload demanding computational processes. After processing these data, control commands can be generated, that shall be sent from the edge back to the robots. Moreover, there is a need for a suitable communication technology as 5G network, to close the entire control loop using an edge architecture. Thus, edge computing will provide high computation power, storage resources and minimal time delays, since edge servers are close to the source of data, while 5G network solutions will reliably provide high data bandwidth with consistent low latencies \cite{ericsson}.

Even with the development of edge computing, some data might need to be sent to the cloud~\cite{wang2020architectural} for further processing and storage purposes. Moreover, the fog computing~\cite{wang2020architectural} paradigm gained high interest, so that the final edge ecosystem for robotic applications could be a hybrid edge-(fog)-cloud robotic system for distributed data and computational processes. A detailed review on edge computing is presented in~\cite{sulieman2022edge}, where researchers focus on edge computing general concepts, while the current work addresses the utilization of edge computing exclusively for robotic applications. 

\begin{figure*}[htbp]
	\centering
	\includegraphics[width=\textwidth]{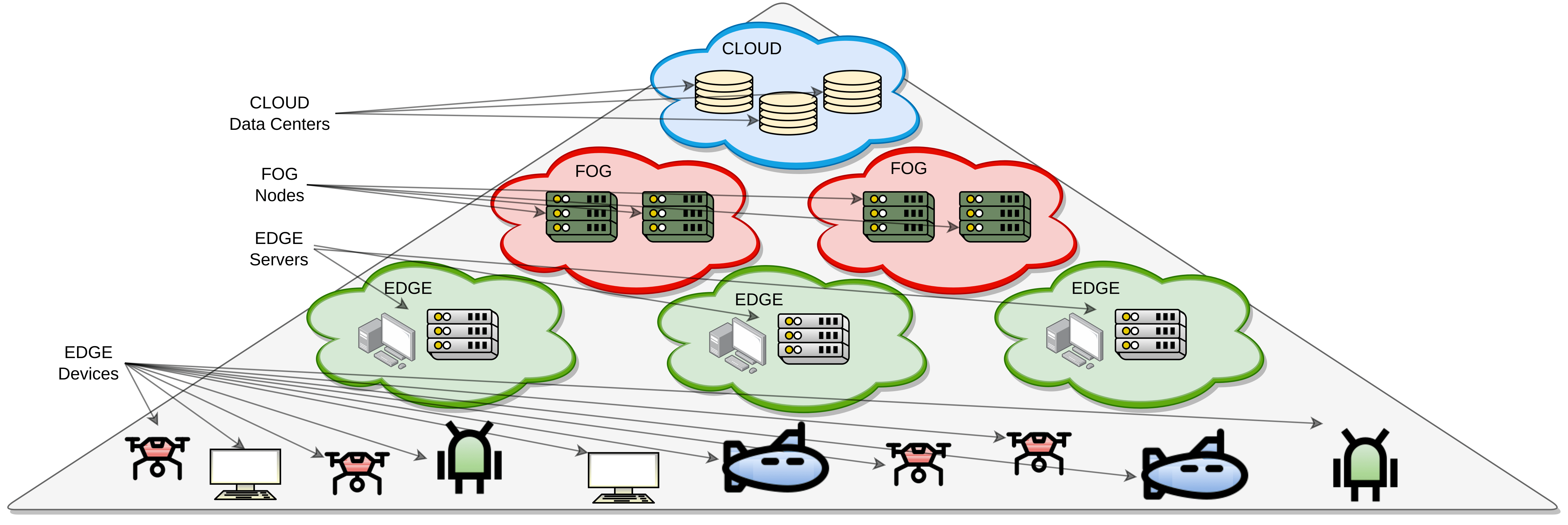}
  	\caption{Data processing and storage layers}
  	\label{fig:layers}
\end{figure*}

Thus, the main contributions of this article is to present, compare and analyze different edge/cloud-based architectures for robotic applications, while analyzing the technologies behind these architectures. Prior to the presentation and analysis of each of the architectures, we will dive into the four layers of computing (CLOUD, FOG, EDGE, ROBOTS) architectures, from a robotic systems perspective and with a focus on what each of these layers offers or enables to the field. Moreover, we will present a brief review of related computational technologies, such as Virtual Machines (VMs), Containers and Kubernetes, why and how each one can be used for deploying applications to the edge. Additionally, simulation and development tools will be presented and compared for a first time ever from a robotics perspective. A comparison of the edge and cloud providers and the types of services they offered, regarding the computational resources for the robotic applications are also examined. To elucidate these terms, in Section ~\ref{apps} we present several edge-based robotic applications and we state the importance of the mentioned approaches.

The rest of the article is structured as it follows. In Section~\ref{clouds} edge/cloud-based architectures for computing robotic systems will be discussed. In Section~\ref{tek} we present the main solutions and software that exist today for edge-based applications, while in Section~\ref{providers} we set forth a review about available edge and cloud providers. In these Sections, it is essential to investigate the advantages and disadvantages of these solutions for each application. In Section~\ref{apps}, we are presenting various robotic applications running mainly on the edge. Finally, we conclude this article in Section~\ref{conclusion} by summarizing all the parameters we have to consider when deciding for a cloud-based solution for robotic applications.
%%%%%%%%%%%%%%%%%%%%%%%%%%%%%%%%%%%%%%%%%%%%%%%%%%%%%%%%%%%%%%%
\section{Cloud - Fog - Edge}
\label{clouds}
%%%%%%%%%%%%%%%%%%%%%%%%%%%%%%%%%%%%%%%%%%%%%%%%%%%%%%%%%%%%%%%
The constantly increasing need for computational power and storage resources has led into the development of various computing architectures~\cite{de2019foundations}. Researchers from several fields of technology like deep learning~\cite{wang2020deep}, video analytics~\cite{wang2022enabling}, industry 4.0~\cite{pace2018edge} are constantly trying to utilize the available architectures or propose new ones, in order to solve existing challenges and to provide new opportunities~\cite{pan2017future}. All these architectures are based on four different computing and storage layers, as depicted in Fig.~\ref{fig:layers}, where the bottom layer is formed from the connected devices. Generally, these layers can be considered as it follows:
\begin{enumerate}
    \item DEVICES LAYER - multiple robotic systems
    \begin{enumerate}
        \item Data generation
        \item Light data processing
        \item Low level control
        \item Other processes needed locally
    \end{enumerate}
    \item EDGE LAYER - offloads computational load from robotic systems to remote entities 
    \begin{enumerate}
        \item Large volume real-time data processing
        \item On premises data visualization
        \item High level control
        \item Micro data storage
    \end{enumerate}
    \item FOG LAYER - extended EDGE LAYER with multiple heterogeneous interconnected nodes
    \begin{enumerate}
        \item Local networking
        \item Control response
    \end{enumerate}
    \item CLOUD LAYER - provides ample storage and computation capabilities, however, alike EDGE and FOG layers, this has long response latency
    \begin{enumerate}
        \item Big data processing
        \item Data storage
    \end{enumerate}
\end{enumerate} 

Giving a closer look at each architectural layer, from the robotics perspective, it will allow us to provide better understanding of the opportunities and challenges that can be addressed using one or another architecture.

At its base, the DEVICES LAYER is represented by individual robots that are equipped with application driven sensors like cameras and lidars for perception, navigation and inspection tasks. These sensors can produce gigabytes of data, which has to be processed on-board in order to enable the autonomous capabilities of the robotic platform. However, due to the limited computational power, only light data processing, which is required, for example, for a single robot SLAM and low level control, can be performed on-board. While in more demanding applications, which involve multiple robots, there is a need to extend this layer with EDGE or FOG layers. These layers allow for real-time data processing of large data batches that are streamed from the network connected robots to the edge server, where heavy computations that require low latencies are carried out. One of such examples is the multi-session SLAM in which multiple SLAM missions are performed collaboratively by multiple robots that explore the same environment. Besides that, the FOG LAYER allows to extend the exploration mission to multiple environments, thus creating a network of heterogeneous interconnected nodes. The CLOUD LAYER excels over the EDGE and FOG layer in terms of storage and computational capabilities, but in comparison to EDGE layer that is located close to the DEVICES layer, CLOUD layer is located much further away from the DEVICE layer. Additionally to that, CLOUD layer is commonly characterized by long response latencies, which can be explained to the long distance between the DEVICE and the CLOUD layer, and thus the robotic challenges for the next generation robotic systems cannot be fully addressed by solely this layer. Thus, one can observe various architectures that are based on the collection of these multiple layers.
%%%%%%%%%%%%%%%%%%%%%%%%%%%%%%%%%%%%%%%%%%%%%%%%%%%%%%%%%%%%%%%
\section{Virtual Machines - Containers - Kubernetes}
\label{tek}
%%%%%%%%%%%%%%%%%%%%%%%%%%%%%%%%%%%%%%%%%%%%%%%%%%%%%%%%%%%%%%%
With the advent of cloud computing and virtualization, many applications apart from robotics are exploring the optimal possible utilization of such technologies. Though virtualization options are a known discussion in the academia and industry, here we make an effort and touch on the three most popular virtualization technologies as well as discuss their characteristics and compare them in terms of benefits and drawbacks for cloud robotics applications. 
\subsection{Virtual Machines}
\begin{figure}[htbp]
	\centering
	\includegraphics[width=\columnwidth]{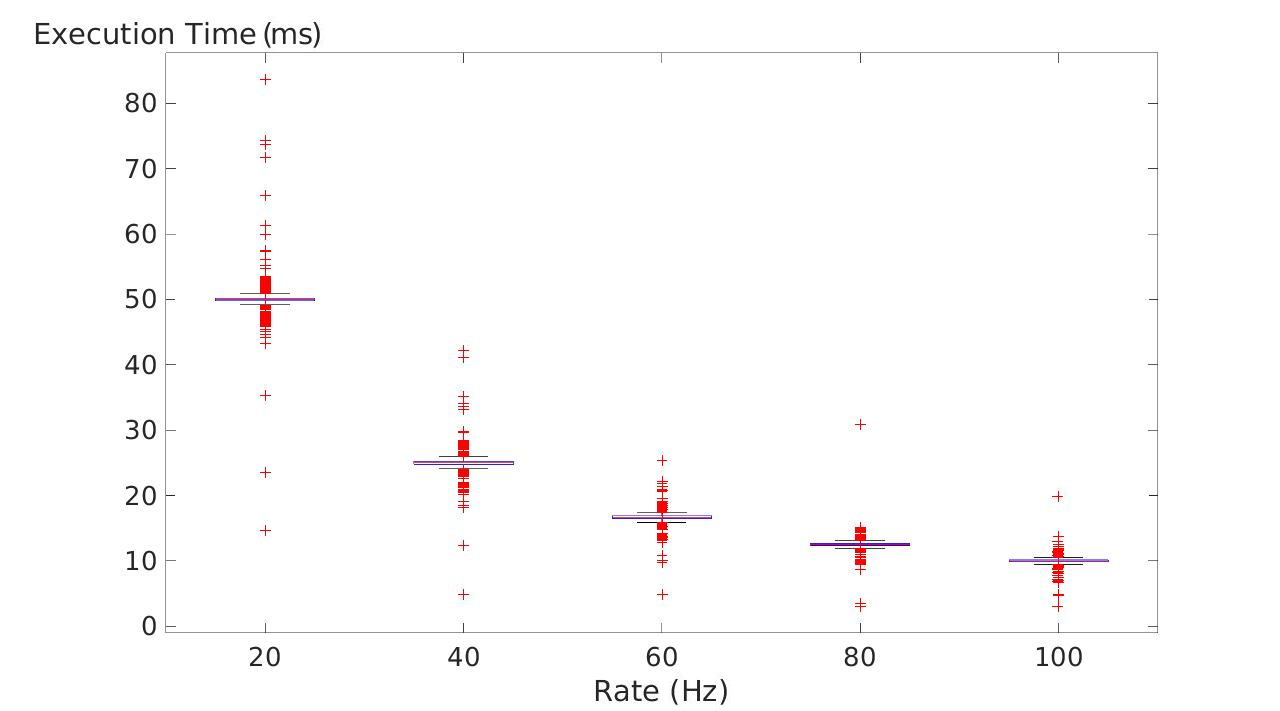}
  	\caption{A procedure was offloaded to the edge as a VM application (also reproduced with containers and k8s). Five different rates where selected for that procedure and the time deviation was measured for each one of them. The mean time of that process coincides with time corresponding rate.}
  	\label{fig:time}
\end{figure}

A virtual machine (VM) is a computer resource that utilizes software instead of a physical computer to deploy applications and run programs and processes \cite{doan2019containers}. VMs also provide a complete emulation of low-level hardware devices like CPU, Disk, RAM, networking devices, and others. VMs can be configured accordingly, to support any application topology and provide a stable platform regarding dependency issues (operating system compatibility, specific software packages, etc.). Specifying computational resources (in Fig \ref{fig:time}, e,g. time deviation of a procedure in the edge with different rates) and resource scaling, redeploying VM applications on different host computers, isolating sensitive applications, as well as running VMs in the Edge cloud are some of the benefits that VMs offer in cloud robotics-related applications. Furthermore, VMs provide a more dynamic and interactive development experience when compared with other virtualization technologies. The latter derives from the fact that VM encompass a full-stack system, as well as utilizing an entire operating system and what one may provide, e.g. a Graphical User Interface. 
\subsection{Containers}
Containers are software packages that include all the software dependencies required to execute the contained software application. The main difference between container and VM is that VM emulates an entire machine down to the hardware layers, while containers only emulate the software components. Such software components might be system libraries, external software packages, and other operating system level applications \cite{pahl2015containers}. Because containers are only emulating software components, they are more lightweight and easy to iterate. Additionally, most container runtime systems provide robust pre-made container (image) repositories. A popular example in robotic applications could be the Robot Operating System (ROS) image. Another key benefit of container technologies is that software packages, contained in the constructed images, can be stacked in levels and produce a novel and more complete application. These flexible characteristics of containers, i.e., lightweight and easy to iterate, are responsible for the birth of another technology, container orchestration.

\begin{figure}[htbp]
	\centering
	\includegraphics[width=\columnwidth]{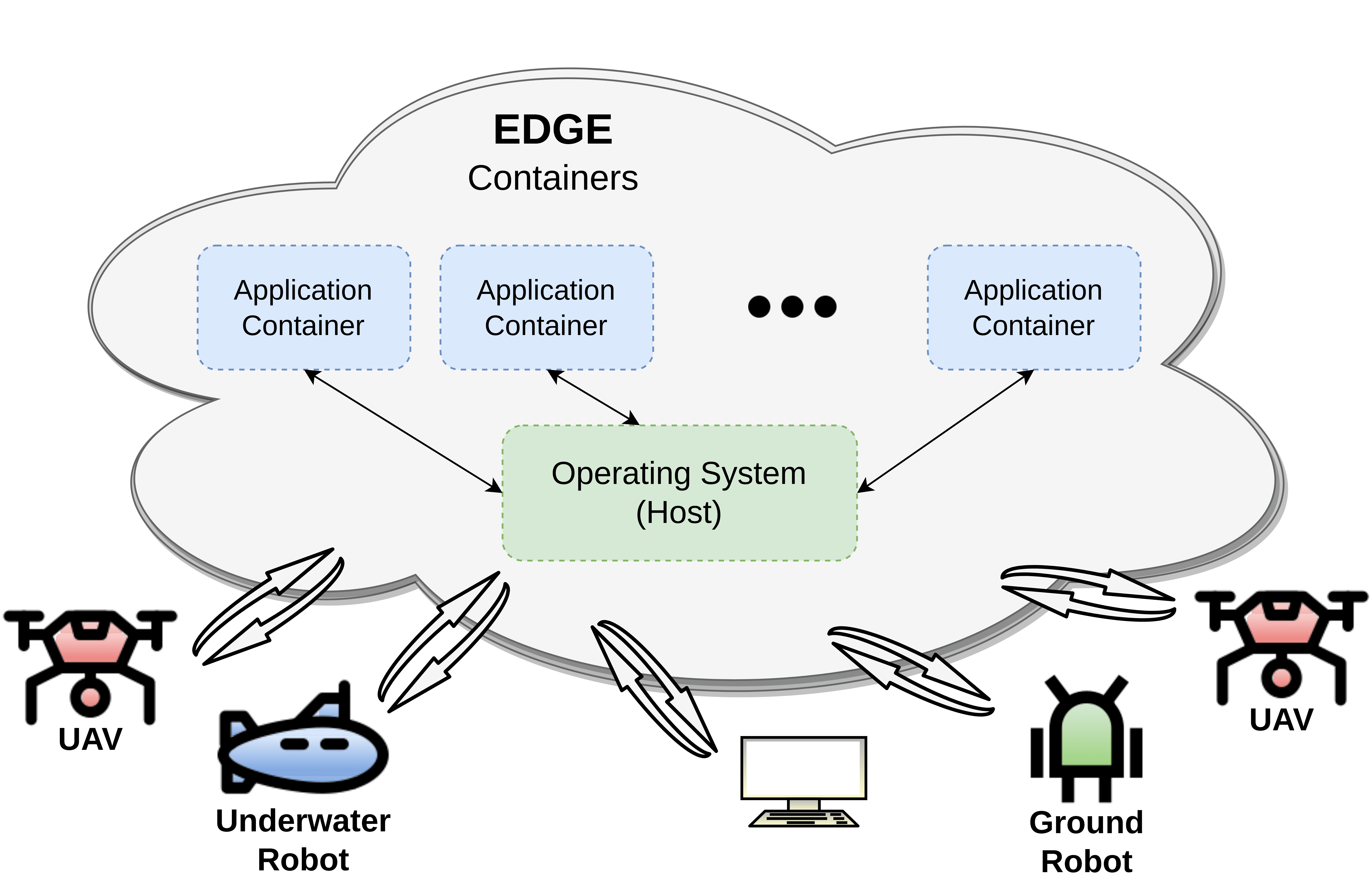}
  	\caption{Architecture of containerized applications for robotic platforms}
  	\label{fig:container}
\end{figure}

The generic topology architecture of containerized applications for robotic systems is shown in Fig \ref{fig:container}. As it is indicated, Data are sent from the robots to the edge and then commands generated from the containers are sent back to robots. Data processing, high level controllers and advanced algorithms can be deployed to the application containers for various robotic missions.

\subsection{Kubernetes}
\begin{figure}[htbp]
	\centering
	\includegraphics[width=\columnwidth]{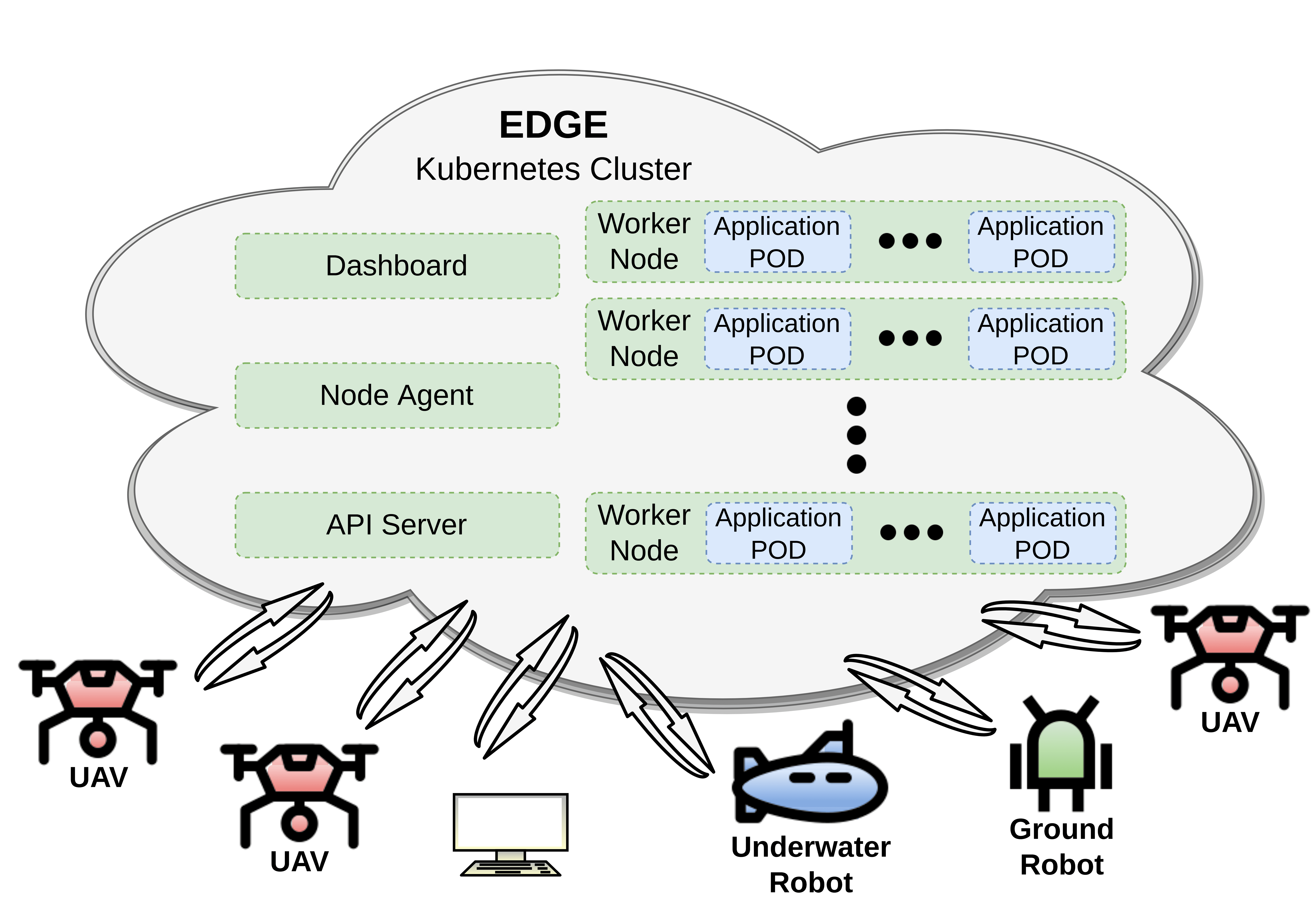}
  	\caption{Kubernetes architecture for robotic platforms}
  	\label{fig:kubernetes}
\end{figure}

Kubernetes (k8s) is an open-source platform that orchestrates container runtime platforms systems across network-connected hardware infrastructure \cite{shah2019building}. K8s was initially developed by Google that needed an efficient way of managing billions of container applications weekly. In such systems, the surface of additional needs became present. Reliability, scalability, robustness, security, and others are essential requirements in real-world applications. k8s bundles a set of containers into a group (called a pod) and manages their life cycle. An example of such a set could be an application server, an SQL database, a Model Predictive Control (MPC) running apart from the robot, a set processing SLAM, and others. K8s will manage these pods performing multiple essential tasks, such as reducing network overhead, increasing resource usage efficiency (load balancing between copies of the same application, either to the same machine or across the cloud cluster), hardware resources designated for your specific configuration (Fig \ref{fig:time}), monitoring and much more since k8s is a rapidly progressing state of-the-art technology. 
To conclude this section we want to present a small topology architecture example illustrated in Fig \ref{fig:kubernetes}. Here you can see how multiple agents are all connected to the Edge cloud where different applications are orchestrated by K8s. In this example, K8s is responsible for scheduling resource-demanding computational jobs (on different worker nodes), managing network traffic and can also be utilized for redundancy requirements.   

%%%%%%%%%%%%%%%%%%%%%%%%%%%%%%%%%%%%%%%%%%%%%%%%%%%%%%%%%%%%%%%
\section{Edge Providers}
\label{providers}

Before we start designing a cloud-based architecture, we have to be aware of the available providers in the related region and the services they provide. In Table \ref{table:cloudsolutions} we go through some of the main cloud and edge providers and cloud solutions that are provided. When we are using the term edge, we are thinking in terms of minimal latency, network hops and distance from the source of data. Thus, it is more likely that local edge providers, that are not mentioned in the Table below, will be better options. Table \ref{table:cloudsolutions} can be utilized as a common ground for acknowledging the available providers and services they offer, while the cloud computing solutions are available in various locations around Europe. Since we are interested in low latency we did not take under consideration solutions that might be available only outside Europe. The sources for Table \ref{table:cloudsolutions} were extracted from~\cite{9673425},~\cite{8938723} and~\cite{9337100}.

\begin{table*}[htbp]
\caption{Comparison overview of the existing cloud solutions for robotic applications}
\label{table:cloudsolutions}
\centering
\begin{tabular}{|c|c|c|c|}
\hline
\textbf{Services and Solutions} & 
\textbf{\begin{tabular}{@{}c@{}}Amazon - AWS\end{tabular}} &
\textbf{\begin{tabular}{@{}c@{}}Microsoft - Azure\end{tabular}} &
\textbf{\begin{tabular}{@{}c@{}}Google - GCP\end{tabular}}
\\ \hline

\textbf{\begin{tabular}{@{}c@{}}Simulation Software for \\ Robotics Application\end{tabular}} & 
\begin{tabular}{@{}c@{}}\textbf{Robomaker:} \\ Good for simulations \\ Deployed applications run on the \\ local robot and not on the cloud \\ Limited location servers\end{tabular} &
\begin{tabular}{@{}c@{}}\end{tabular} & 
\begin{tabular}{@{}c@{}}\end{tabular}
\\ \hline

\textbf{\begin{tabular}{@{}c@{}}Virtual Machines\end{tabular}} & 
\begin{tabular}{@{}c@{}}\textbf{Elastic Compute Cloud (EC2):} \\ Access other AWS services \\ Many OS options \\GPU graphic or compute instances \\ Security \\ Many location servers\\ Complex \\ Price variations (expensive) \\ \textbf{Lightsail:} \\ Simple and easy to use \\ Upgradable to EC2 \\ Affordable \\ Images on virtual private server \\ Instances of several OS \\ Some location servers \\ Some service options \\ Limited CPU capacity\end{tabular} &  
\begin{tabular}{@{}c@{}}\textbf{Azure Virtual Machine (AVM):} \\ High availability \\ Security \\ Scalability \\ Slow \\ Few options for low size VMs\end{tabular} &  
\begin{tabular}{@{}c@{}}\textbf{Google Compute Engine (GCE):} \\ Easy to use \\ Many CPU and GPU options \\ High-performance VMs \\ Easy to scale \\ Container support \\ Some location servers \\ High overhead \\ Slow auto-scaling\end{tabular} 
\\ \hline

\textbf{\begin{tabular}{@{}c@{}}Containers\end{tabular}} & 
\begin{tabular}{@{}c@{}}\textbf{Elastic Container Service (ECS):} \\ Manages containers in clusters \\ Integrates with the AWS ecosystem \\ Control plane is totally free \\ Many location servers\end{tabular} & 
\begin{tabular}{@{}c@{}}\textbf{Azure Container Instances (ACI):} \\ Instances of Linux and Microsoft OS \\ Many CPU and GPU options \\ Many location servers \\ Fast and easy to use \\ Security \\ Complex \\ Expensive\end{tabular} &  
\begin{tabular}{@{}c@{}}\textbf{} \\ \end{tabular} 
\\ \hline

\textbf{\begin{tabular}{@{}c@{}}Kubernetes\end{tabular}} & 
\begin{tabular}{@{}c@{}}\textbf{Elastic Kubernetes Service (EKS):} \\ Flexible \\ Easy to migrate workload to another \\ platform \\ Good for complex applications \\ Offers more control \\ Control plane is charged\end{tabular} & 
\begin{tabular}{@{}c@{}}\textbf{Azure Kubernetes Service (AKS):} \\Flexible \\ Reduced management overhead \\ Control Panel is free \\ Some location servers\end{tabular} & 
\begin{tabular}{@{}c@{}}\textbf{Google Kubernetes Engine (GKE):} \\ Open-source container orchestration \\ Prebuilt kubernetes applications and \\ templates \\ Ease to migrate traditional workloads \\ Limited location servers\end{tabular}
\\ \hline

\end{tabular}
\end{table*}

These cloud solutions, can work as an ecosystem. For example, Amazon Web Services (AWS)~\cite{forstermasters1998} can be combined so Amazon Elastic Kubernetes Service (EKS) can be integrated to Amazon Elastic Compute Cloud (EC2), to deploy and manage containerized applications at scale. The same idea applies with Google Cloud~\cite{google}, where Kubernetes Engine (GKE) monitoring tools can be integrated to Google Cloud Compute Engine (GCE).

\subsection{Amazon AWS}
Amazon's AWS provide several options in terms of cloud computing and is the leader of cloud computing. Each option has some advantages and disadvantages. We will go through the Table's \ref{table:cloudsolutions} data and analyze the options for deploying a robotic application. Amazon AWS Robomaker is a good cloud-based option for developing, testing, deploying intelligent robotics applications and running simulations. The downside is that there are not many servers across the world so in most cases Robomaker can not be used as Edge. An additional drawback and reason that Robomaker can not be used as Edge is that although the development happens on cloud, the deployed application runs locally on the robot. Amazon Lightsail is an affordable and good option for deploying applications with low demands. It is easy to deploy instances of several OS and has the option to be upgraded to EC2 for higher demands. EC2 is the main cloud compute option of AWS. It offers plenty OS options and there are available servers all across the world, and provides the choice for Kubernetes orchestration for the containerized applications~\cite{forstermasters1998},~\cite{bartwal_dhyani_subedi_dangayach_2021}.

\subsection{Microsoft Azure}
Microsoft Azure has less compute services compared to AWS and less location servers and is the second largest platform for cloud/edge computing solutions. Azure offers smooth hybrid-cloud environment which is of great importance as mentioned in Section~\ref{clouds}. AVMs provide security and scalability but few options for low size VMs. In contrast, Azure ACI and AKS provide a wide range of solutions, such as GPU, CPU options, fast and easy environment and are available in many locations~\cite{directoryofazure},~\cite{bartwal_dhyani_subedi_dangayach_2021}.

\subsection{Google Cloud}
Google Cloud offer less location and service options compared to AWS, but has a handful of options for cloud solutions. GCE provide many CPU and CPU option, high performance VMs and container support. The GKE offers prebuild kubernetes applications and templates and it is quite easy to migrate traditional workloads. Additionally provides outstanding open-source options~\cite{google},~\cite{bartwal_dhyani_subedi_dangayach_2021}.

Latency, safety concerns are some of the issues that should be addressed. The continuous development and the combination of edge computing and 5G networks can be the answer to these challenges. When thinking of edge computing solutions, since time delays is one of the most important parameters, local edge providers should be considered. Local providers, in most cases, will have their servers closer to the client's source of generated data, thus they will be able to provide lower latency. Additionally, requests can be more flexible regarding provided services. Most of the times, local providers offer VMs, docker hub and Kubernetes solutions as well as several storage and security options. 
%%%%%%%%%%%%%%%%%%%%%%%%%%%%%%%%%%%%%%%%%%%%%%%%%%%%%%%%%%%%%%%
\section{Edge Robotics Applications}
\label{apps}
In this Section, we present some edge-based robotic applications, selected from research articles and the state-of-art. The following applications, use some of the technologies, architectures and services mentioned above.

Computational offloading to the edge and or the cloud for robotic platforms is a topic that nowadays has become more and more relevant. The generated amount of data that robots produce and has to be analyzed at each moment in almost real-time, must be respected. Offloading this computational effort will bring great advantages in terms of latency and computational capacity, which is a very important factor to keep in mind. Researchers and engineers have been trying, in the last couple of years, to offload some computational demanding procedures to the edge and the results in some cases are notable.

In~\cite{9090991}, a platform is presented for a heterogeneous robotic system focused on search missions. A lightweight traversal algorithm deployed in the edge is proposed for high available search. A group of robots will perform a cooperative mission to search and find artifacts in a predefined area, quickly and intelligently, while edge computing is used for offloading the task partitioning algorithm, considering real data processing and low latency. Thus, the UAV sends data to the edge when performing traversal tasks. In~\cite{gudi2018fog}, social fog robotics systems are introduced for responsive human-robot interaction and several architectures are presented. The response rate of robots is validated and the whole system is examined in terms of latency. SLAM tasks are offloaded to the edge and the system's architecture is presented in~\cite{sarker2019offloading}, where edge is used for real-time decisions. The architecture of this system has the following four layers, each one for a different cause. The robot and edge layer is used for data processing and analysis, while the fog layer is used for distributed storage and the cloud layer is used for monitoring and for general mission control. In~\cite{sun2021cloud} the edge is used to reduce the delay and energy consumption of deep learning models by analyze the performance of edge servers and estimate time delays, to establish a mechanism on intelligent data envelopment for intelligent social robots. In~\cite{tanwani2019fog}, deep robot learning distributes compute, storage and networking resources between the cloud and the edge for the task of object recognition and grasp planning by a mobile robot. In~\cite{tian2019fog}, the edge is used for offloading a dynamic self-balancing navigation controller, while cloud is used for assisted teleoperation and visual recognition of dynamic self-balancing robots that can detect and pick up objects, while being teleoperated.

Some articles are focused on offloading controllers on the edge. In~\cite{aarzen2018control},~\cite{skarin2019cloud} and~\cite{skarin2020cloud2} the MPC is proposed to be offloaded to the edge. These articles are focused mainly on evaluating latency, long time delays and uncertainty. The control in~\cite{skarin2019cloud} is composed with the combination of a LQR (locally) and MPC (Edge) control, while in~\cite{aarzen2018control} two MPC are implemented, one on a local Edge and one on a Cloud. The computationally heavy MPC is offloaded to the Edge. In~\cite{skarin2020cloud2} a variable horizon strategy for a cloud-based MPC is presented and a remote MPC is used to control a ball and bean system. In~\cite{tsokalo2019mobile} applications, containing remote controller, run in a Mobile Edge server in a form of Docker Containers. These application are controlling two robotic arms for cooperation tasks in an industrial environment, while in~\cite{salman2019fogification} and in~\cite{shaik2020enabling} the utilization of fog layer to enhance the capabilities of industrial robotics has been proposed. In~\cite{ma2020exploring} a switching multi-tier control is presented. The switcher is choosing between a local controller which operates as a safety controller and an Edge controller which runs more sophisticated algorithms with optimal performance.

As mentioned in Section~\ref{tek}, in many cases edge computing is based on kubernetes components. An edge architecture and an open-source network is presented in~\cite{figueiredo2020edgevpn}, where the kubernetes container orchestration middleware is highlighted for the deployment of virtual networks spanning distributed edge and cloud resources. The goal of~\cite{kochovski2019architecture} is to automate the process of making decision in terms of placement of the expected workload. In this article, generated data from various sensors from smart home, cities, construction and robots are distributed to edge, fog and cloud. This process is achieved by using kubernetes orchestration. In~\cite{cha2021design}, studied plugins based on Container Network Interface (CNI) to provide low-latency edge services, that can be used in robotic applications \cite{qi2020understanding}. Finally, in~\cite{lumpp2021container}, a methodology based on docker and kubernetes for ROS-based robotic application is presented. In that case, the architecture was evaluated by experimental results obtained by a mobile robot interacting with an industrial agile production chain.

Mobile edge computing (MEC) has been recognized as a promising technique. Thank to the advantages of saving energy for users, providing low latency and achieving security, MEC has received increasing attention in both academia and industry. In that context, a different but interesting approach is presented in~\cite{zhou2018uav}. In this article, researcher suggested a MEC system, where aerial robots are equipped with servers to provide MEC services to ground users.
%%%%%%%%%%%%%%%%%%%%%%%%%%%%%%%%%%%%%%%%%%%%%%%%%%%%%%%%%%%%%%%
\section{Conclusions}
\label{conclusion}
Edge-based robots set the state-of-art in the field of robotics. The significant possibilities have attracted the attention of researchers and engineers. Despite the attention, there is great room for improvement and future directions. Even though edge can play a remarkable role in multi-agent systems, in terms of robot communication and collaboration, there are not so many articles towards this direction. Furthermore, edge capabilities regarding computational power can be be exploited even more in the future, so more heavy and demanding tasks can be offloaded there. The rapid development of communication (5G networks) and computing technologies (cloud, fog, edge computing) will provide more possibilities and solve current challenges, hence will help the robotics field to move towards fully autonomous systems. In that concept, we investigated, presented, analyzed and compared the different architectures, typologies and their components. Most of these technologies and methods are recently developed thus, we wanted to display a general idea of all these terms and spread awareness of various solutions. Additionally, we would like to inspire researchers to engage with this area and contribute in their own way.
%%%%%%%%%%%%%%%%%%%%%%%%%%%%%%%%%%%%%%%%%%%%%%%%%%%%%%%%%%%%%%
\bibliographystyle{./IEEEtranBST/IEEEtran}
\bibliography{./IEEEtranBST/IEEEabrv,references}

%%%%%%%%%%%%%%%%%%%%%%%%%%%%%%%%%%%%%%%%%%%%%%%%%%%%%%%%%%%%%%
\end{document}